\documentclass[11pt]{article}

\usepackage{cite}
\usepackage{amsmath,amssymb,amsfonts}
\usepackage{algorithmic}
\usepackage{graphicx}
\usepackage{textcomp}
\usepackage{subcaption}
\usepackage{xcolor}
\usepackage{booktabs}
\usepackage{multirow}
\usepackage{makecell}
\usepackage{comment}

\begin{document}

\title{Hand Gesture Recognition from Doppler Radar Signals Using Echo State Networks}

\author{Towa Sano$^{1}$ and Gouhei Tanaka$^{1,2}$\\
\small $^{1}$Department of Computer Science, Nagoya Institute of Technology, \\Nagoya 466-8555, Japan\\
\small $^{2}$International Research Center for Neurointelligence, \\The University of Tokyo, Tokyo 113-0033, Japan\\
\small \texttt{t.sano.629@stn.nitech.ac.jp}, \texttt{gtanaka@nitech.ac.jp}
}

\date{}

\maketitle

\begin{abstract}
Hand gesture recognition (HGR) is a fundamental technology in human computer interaction (HCI).
In particular, HGR based on Doppler radar signals is suited for in-vehicle interfaces and robotic systems, necessitating lightweight and computationally efficient recognition techniques.
However, conventional deep learning-based methods still suffer from high computational costs.
To address this issue, we propose an Echo State Network (ESN) approach for radar-based HGR, using frequency-modulated-continuous-wave (FMCW) radar signals.
Raw radar data is first converted into feature maps, such as range-time and Doppler-time maps, which are then fed into one or more recurrent neural network-based reservoirs. 
The obtained reservoir states are processed by readout classifiers, including ridge regression, support vector machines, and random forests.
Comparative experiments demonstrate that our method outperforms existing approaches on an 11-class HGR task using the Soli dataset and surpasses existing deep learning models on a 4-class HGR task using the Dop-NET dataset.
The results indicate that parallel processing using multi-reservoir ESNs are effective for recognizing temporal patterns from the multiple different feature maps in the time-space and time-frequency domains. 
Our ESN approaches achieve high recognition performance with low computational cost in HGR, showing great potential for more advanced HCI technologies, especially in resource-constrained environments.

\end{abstract}


\section{Introduction}
Hand gesture recognition (HGR) is a fundamental technology for realizing intuitive and natural human-computer interaction (HCI) by utilizing human body movements as input signals. Vision-based HGR methods using RGB or depth cameras have achieved remarkable recognition accuracy through advances in deep learning techniques \cite{molchanov2016online}. However, these approaches face several practical challenges that limit their deployment in real-world scenarios. First, camera-based systems are sensitive to lighting conditions, with performance degrading significantly in low-light or backlit environments. Second, the operating range is constrained by camera placement and field of view\cite{ahuja2021vid2doppler}.

Radar-based sensing has emerged as a promising alternative that addresses these fundamental limitations. 
Radar systems measure distance and velocity using electromagnetic waves, enabling robust operation regardless of ambient lighting conditions \cite{Ahmed2021hand, Soumya2023Review}. 
Among various radar technologies, frequency-modulated-continuos-wave (FMCW) millimeter-wave radar, providing Doppler information, is particularly attractive for gesture sensing. 
Operating in the high frequency band, FMCW radar can simultaneously acquire high-resolution range and velocity information with compact form factors suitable for integration into consumer electronics \cite{jankiraman2018fmcw, Skaria2023MmWaveReview}.

Deep learning models achieve high accuracy in HGR tasks using FMCW radar signals, but their high computational cost makes them difficult to deploy on edge devices requiring real-time performance.
To address the computational efficiency challenge while maintaining high classification performance, we propose an Echo State Network (ESN)-based approach for HGR using Doppler radar signals. 
ESN is a typical reservoir computing (RC) model that employs a recurrent neural network (RNN) with fixed random internal weights as a reservoir, requiring only the training of output layer weights through simple algorithms \cite{jaeger2001echo}. This architecture eliminates the need for computationally expensive backpropagation through time, significantly reducing training and inference costs while maintaining the ability to capture complex temporal dependencies in sequential data \cite{tanaka2019reservoir}.

In our method, raw radar signals corresponding to hand gestures are transformed into feature maps in the time–space and time–frequency domains, which are then processed with an ESN-based approach. In a conventional single-reservoir ESN, all feature maps are concatenated and fed into a single reservoir. However, integrating heterogeneous information in this manner may cause interference within the reservoir. To address this issue, we propose an approach in which different feature maps are independently input to multiple reservoirs, and the resulting reservoir states are integrated and fed into a readout layer. This approach is inspired by multi-reservoir computing models \cite{gallicchio2017deep,li2023multi}. For training the readout layer, we employ not only standard ridge regression but also support vector machines (SVMs) and random forests.

To evaluate the effectiveness of the proposed approach, we perform an 11-class HGR task using range–time and Doppler–time maps derived from the Soli dataset \cite{wang2016interacting}. The results demonstrate that the proposed multi-reservoir approach outperforms existing methods while achieving shorter training and inference times than deep learning–based baselines. Furthermore, we also report results on a 4-class HGR task using the Dop-NET dataset \cite{RitchieDopNet}, to which a single-reservoir ESN is applied because only a single feature map (micro-Doppler map) is provided.

Our key contributions are summarized as follows:
\begin{itemize}
\item  This is the first work to apply an ESN-based approach to radar-based HGR, to the best of our knowledge. Moreover, we propose a parallel multi-reservoir ESN architecture to mitigate the interference of multiple different feature maps.
\item Experimental results demonstrate that the proposed approach achieves 98.84\% classification accuracy and outperforms existing methods on the Soli dataset, while operating with much shorter inference times than deep-learning-based approaches.
\item The performance of the proposed approach is comprehensively evaluated using cross-validation, leave-one-subject-out, and leave-one-session-out evaluation settings.
\end{itemize}

\section{Related Work}
\subsection{Radar-Based Hand Gesture Recognition}
Radar-based HGR has evolved significantly in the past decade, driven by advances in both hardware miniaturization and machine learning algorithms. 
Early approaches relied on continuous-wave radar, which provides velocity information but lacks range discrimination. 
FMCW radar addresses this limitation by enabling simultaneous range and velocity estimation, making it the predominant choice for modern gesture sensing systems \cite{Soumya2023Review, Skaria2023MmWaveReview}.

Project Soli, developed by Google ATAP, pioneered the use of 60 GHz FMCW millmeter-wave radar for fine-grained gesture sensing \cite{Lien2016Soli}.
The system employs a compact radar module with one transmit and four receive antennas, capable of detecting sub-millimeter movements.
The original Soli work employed statistical features such as mean, variance, peak positions extracted from range-Doppler maps (RDMs), combined with random forest classifiers optimized for embedded deployment. 
This approach achieved 92.10\% gesture-level accuracy on 5-class recognition with Bayesian post-processing for temporal smoothing.

Wang et al. \cite{wang2016interacting} systematically evaluated feature engineering–based gesture recognition pipelines on the Soli dataset and showed that performance saturates when relying on hand-crafted, frame-level representations. 
Using random forest classifiers with per-frame features achieved only 26.8\% accuracy for 10-class recognition, and incorporating simple temporal difference features improved accuracy marginally to 30.0\%. 
Hidden Markov Models, which explicitly model temporal transitions, reached 35.0\%, indicating that limited temporal modeling alone is insufficient under hand-designed feature representations. 
To overcome these limitations, the authors proposed an end-to-end CNN–LSTM architecture that jointly learns discriminative representations from range-Doppler sequences and captures temporal dynamics through recurrent modeling. 
This approach achieved 87.17\% frame-level accuracy, further improving to 94.15\% in a leave-one-session-out protocol.

Zhang et al. \cite{Zhang2022ResNetLSTM} improved on the CNN-LSTM framework by replacing the convolutional backbone with ResNet-18. The residual connections and batch normalization enabled stable training on the relatively low-resolution $32 \times 32$ RDMs, achieving 92.55\% accuracy. However, like other deep learning approaches, this method requires substantial GPU resources for training and may face deployment challenges on resource-constrained devices.

\subsection{Reservoir Computing Approaches}
RC is a computational framework consisting of a fixed {\it reservoir} for mapping sequential inputs into a high-dimensional feature space and a trainable {\it readout} for mapping reservoir states into a desired output. 
ESN is the most widely adopted RC model for continuous-valued signals, derived from RNN \cite{jaeger2001echo}.
ESN has demonstrated competitive performance across diverse time-series tasks including speech recognition, financial prediction, and biomedical signal processing \cite{yan2024emerging}. 
The key insight is that a sufficiently large and appropriately configured reservoir can project input sequences into a high-dimensional state space where temporal patterns become linearly separable.
Since ESN has an advantage in its computational efficiency compared with fully trainable RNNs, this approach is promising for efficient radar-based HGR.
However, to the best of our knowledge, no prior work has attempted it.

Tsang et al. \cite{tsang2021radar} applied Liquid State Machines (LSM) \cite{Maass2002RealTime}, which employs a spiking neural network-based reservoir in the RC framework, to radar-based HGR. Their approach converts RDM sequences to spike trains through threshold-based encoding, processes them through a reservoir of approximately 1,000 spiking neurons, and classifies the resulting state vectors using various readout methods. With optimized hyperparameters, this approach achieved an accuracy of 98.02\% when using SVM readout under 50:50 holdout split on the Soli dataset, and 98.6\% under 10-fold cross-validation. On the Dop-NET dataset, it achieved 98.91\% when using SVM readout under 50:50 holdout splitting, and 98.6\% under 10-fold cross-validation.
However, the spike encoding preprocessing adds computational overhead, and requires careful parameter tuning to achieve appropriate firing activity in the reservoir.

\section{Proposed Method}
\subsection{Feature Extraction}
FMCW radar transmits linear frequency-modulated signals, known as chirps. 
The received signal is mixed with the transmitted signal to generate an intermediate frequency signal. 
The RDM is then constructed through the following two-stage fast Fourier transform (FFT) process:
\begin{itemize}
    \item Range FFT: An FFT is applied to each individual chirp (fast-time axis) to extract the beat frequency, which corresponds to the target's distance.
    \item Doppler FFT: A second FFT is performed across a sequence of consecutive chirps (slow-time axis). This detects phase shifts between chirps caused by the Doppler effect, revealing the target's radial velocity.
\end{itemize}

The resulting RDM is a two-dimensional intensity map representing the reflection energy in the range-velocity domain. Since dynamic gestures produce non-stationary signals where both range and velocity components vary over time, we decompose the RDM sequence into the range-time map (RTM) and the Doppler-time map (DTM). This transformation enables the model to capture modality-specific temporal dynamics while significantly reducing the input dimensionality for efficient processing.

RTM aggregates velocity information to emphasize range dynamics as follows:
\begin{equation}
\text{RTM}(r,t) = \sum_{d} \text{RDM}(r,d,t) ,
\end{equation}
where $r$, $d$, and $t$ denote the range bin index, Doppler bin index, and time frame index, respectively.

DTM aggregates range information to emphasize velocity dynamics as follows:
\begin{equation}
\text{DTM}(d,t) = \sum_{r} \text{RDM}(r,d,t).
\end{equation}

\begin{figure}[t]
  \centering
  \begin{subfigure}{0.32\linewidth}
    \centering
    \includegraphics[width=\linewidth]{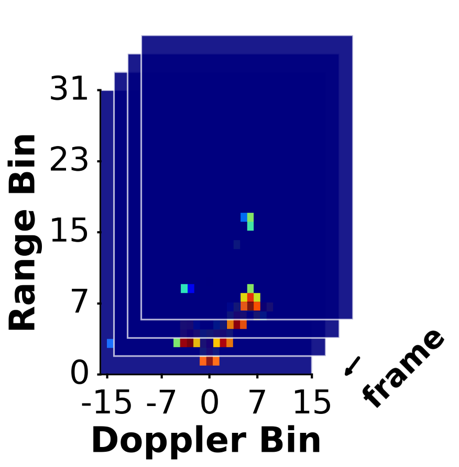}
    \caption{RDM}
    \label{fig:rdm}
  \end{subfigure}
  \hfill
  \begin{subfigure}{0.32\linewidth}
    \centering
    \includegraphics[width=\linewidth]{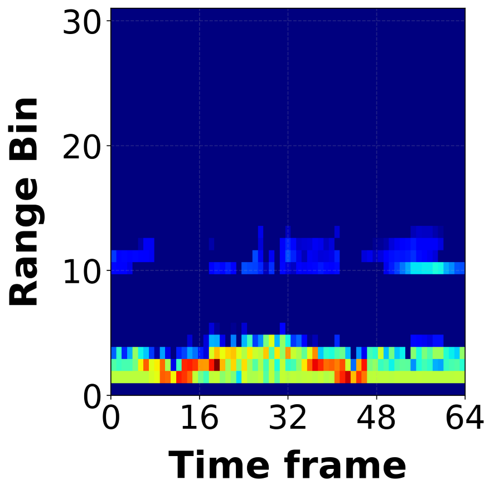}
    \caption{RTM}
    \label{fig:rtm}
  \end{subfigure}
  \hfill
  \begin{subfigure}{0.32\linewidth}
    \centering
    \includegraphics[width=\linewidth]{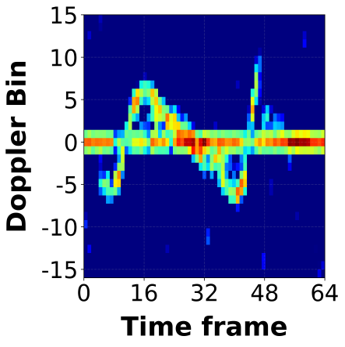}
    \caption{DTM}
    \label{fig:dtm}
  \end{subfigure}
  \caption{Examples of feature maps: (a) range-Doppler map (RDM), (b) range-time map (RTM) and (c) Doppler-time map (DTM).}
  \label{fig:radar_maps}
\end{figure}

Fig. \ref{fig:radar_maps} illustrates examples of the feature maps obtained from raw radar signals. 
Fig. \ref{fig:radar_maps}(a) shows an example of RDM. Figs. \ref{fig:radar_maps}(b) and \ref{fig:radar_maps}(c) show examples of the RTM and DTM, respectively, which are generated by processing the RDM in Fig. \ref{fig:radar_maps}(a) according to the aggregation method described above.
The Doppler profile is centered at zero, corresponding to no movement, whereas positive or negative values indicate motion towards or away from the radar receiver. 

The Soli dataset provides RDMs. We generate the corresponding RTMs and DTMs from the RDMs for each of the four receive antennas in our numerical experiments.

\begin{figure}[t]
\centering
\includegraphics[width=1\columnwidth]{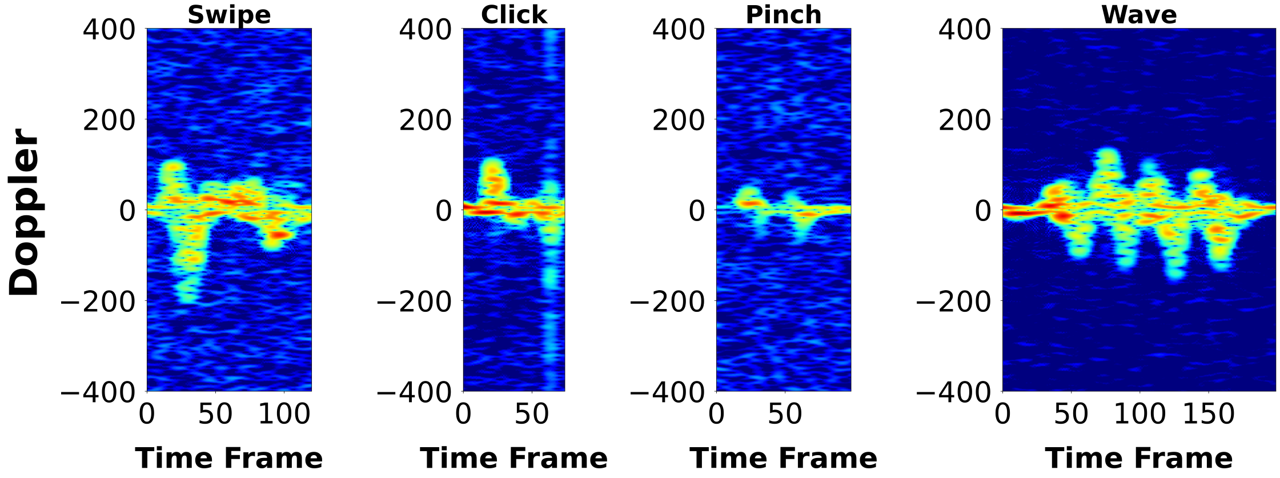}
\caption{Examples of micro-Doppler map (MDM).}
\label{fig:Dop-NET}
\end{figure}

For the Dop-NET dataset, which provides only micro-Doppler maps (MDMs) instead of RDMs, we use the MDMs as feature maps fed into the subsequent classifier. MDMs are obtained by applying short-time Fourier transform (STFT) to radar signals along the slow-time dimension, capturing fine-grained motion-induced Doppler variations over time.
Fig. \ref{fig:Dop-NET} illustrates examples of MDMs for the four types of gestures.

We note that our method, which relies on the systematic feature extraction described above, maintains computational simplicity without requiring additional preprocessing steps such as image resizing, background subtraction, or statistical feature extraction.

\subsection{Echo State Network (ESN)}
\label{sec:ESNs}
\begin{figure}[!b]
\centering
\includegraphics[width=0.9\columnwidth]{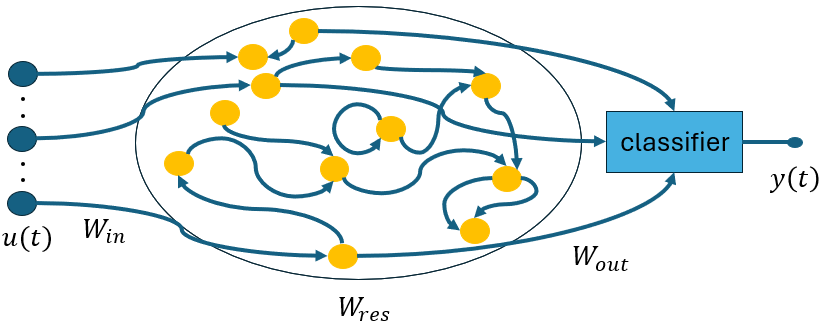}
\caption{Echo State Network architecture. The reservoir layer contains recurrently connected nodes with fixed random weights. Only the output layer weights $\mathbf{W}_{out}$ are trained.}
\label{fig:esn}
\end{figure}

An ESN consists of three components, which are an input layer, a reservoir layer, and an output layer as illustrated in Fig. \ref{fig:esn}. The input layer projects the input signal $\mathbf{u}(t) \in \mathbb{R}^{N_{in}}$ into the reservoir through a fixed weight matrix $\mathbf{W}_{in} \in \mathbb{R}^{N \times N_{in}}$, where $N$ denotes the number of reservoir nodes.

The reservoir layer consists of $N$ recurrently connected nodes with fixed weight matrix $\mathbf{W}_{res} \in \mathbb{R}^{N \times N}$. The state update equation with leaky integration is described as follows \cite{jaeger2007optimization}:
\begin{equation}
\mathbf{x}(t+1) = (1-\alpha)\mathbf{x}(t) + \alpha f\left(\mathbf{W}_{in}\mathbf{u}(t+1) + \mathbf{W}_{res}\mathbf{x}(t) \right),
\label{eq:esn}
\end{equation}
where $\mathbf{x}(t) \in \mathbb{R}^N$ is the reservoir state vector,  $f(\cdot) = \tanh(\cdot)$ is the activation function, and $\alpha \in (0,1]$ is the leaking rate that controls the trade-off between memory retention and input responsiveness.
The spectral radius $\rho = \max|{\mathrm{eigs}}(\mathbf{W}_{res})|$ is a critical hyperparameter.
To ensure that reservoir dynamics depend only on input history after a transient period, $\rho$ is typically scaled to values near or slightly below 1.0. Larger $\rho$ values enable longer memory but risk instability.

When we use a linear readout, the model output is given by the weighted sum of the reservoir states as follows:
\begin{equation}
\label{eq:y}
\mathbf{y}(t+1) = \mathbf{W}_{out}\mathbf{x}(t+1),
\end{equation}
where $\mathbf{W}_{out}$ is the only trainable output weight matrix.
For classification tasks, generally, we use the final reservoir state after processing the complete input sequence, and train $\mathbf{W}_{out}$ using ridge regression with regularization parameter $\lambda$ to prevent overfitting.
The optimal output weight matrix $\mathbf{W}_{out}$ is obtained by minimizing a regularized least-squares objective, which admits a closed-form solution given by
\begin{equation}
\mathbf{W}_{out} = \mathbf{Y}\mathbf{X}^\top \left( \mathbf{X}\mathbf{X}^\top + \lambda \mathbf{I} \right)^{-1},
\end{equation}
where $\mathbf{X}$ denotes the matrix of collected reservoir states and $\mathbf{Y}$ represents the corresponding target labels.

In addition to the linear readout trained by ridge regression, we also test nonlinear readouts as described in Sec. \ref{sec:ESN}. 

\subsection{Multi-Reservoir ESN}
The Soli dataset provides radar signals collected from four receive antennas, producing four-channel RDM sequences. Converting each channel to RTM and DTM yields eight distinct feature maps per sample. These inputs possess inherently different characteristics because RTM emphasizes spatial range dynamics, whereas DTM captures velocity patterns, and each antenna channel observes the gesture from a slightly different spatial perspective.

\begin{figure}[t]
\centering
\includegraphics[width=\columnwidth]{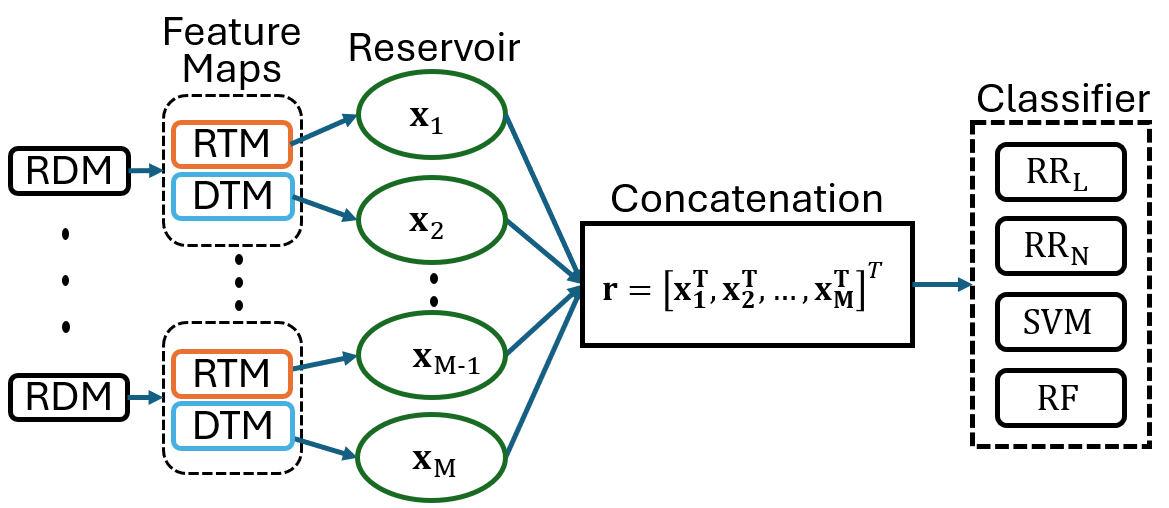}
\caption{Multi-reservoir ESN architecture. The $M$ independent reservoirs process RTMs and DTMs obtained from multiple antenna channels. Final states of all the reservoirs are concatenated and then fed into a readout classifier.}
\label{fig:multi_res}
\end{figure}

Feeding all eight maps into a single reservoir in the ESN may cause interference between modalities, since the shared reservoir dynamics must simultaneously encode heterogeneous signal characteristics.
To address this issue, we adopt a multi-reservoir ESN architecture illustrated in Fig. \ref{fig:multi_res}, inspired by grouped ESNs and parallel multi-reservoir architectures \cite{gallicchio2017deep,li2023multi}.

Each input feature map is processed by an independent reservoir consisting of a relatively small number of nodes. All reservoirs of the same size operate in parallel, allowing each reservoir to learn modality-specific dynamics. Once the input sequence has been fully processed, the final state vectors of all reservoirs are concatenated to construct a unified reservoir state given by
\begin{equation}
\mathbf{r} = [\mathbf{x}_1^T, \mathbf{x}_2^T, \ldots, \mathbf{x}_M^T]^T \in \mathbb{R}^{M N},
\end{equation}
where $N$ represents the number of nodes in each reservoir and $M$ indicates the number of feature maps. This concatenated vector $\mathbf{r}$ is used as the input to the readout (see Sec. \ref{sec:clas}).

This architecture provides several advantages. 
First, modality-specific temporal features can be preserved while minimizing cross-interference between heterogeneous inputs. 
Second, each reservoir can be configured to better accommodate the characteristics of its corresponding input modality. 
Third, the use of relatively small individual reservoirs reduces the computational cost per reservoir, while the parallel combination yields a rich and expressive feature representation.

\subsection{Readout Classifier}
\label{sec:clas}
We test the following four readout classifiers.
\begin{itemize}
\item Ridge Regression (RR$_{\mathrm{L}}$): A linear classifier with $L^2$ regularization, where the regularization coefficient $\lambda$ is set to 0.1. This classifier admits a closed-form solution, enabling fast and efficient training as described in Sec. \ref{sec:ESNs}.
\item Ridge Regression with Nonlinear Transformations (RR$_{\mathrm{N}}$): Ridge regression applied after a nonlinear feature expansion to the unified reservoir state. The feature expansion is represented as follows:
\begin{equation}
\label{eq:multiRes}
\boldsymbol{\Psi}(\mathbf{r}) = [1, \mathbf{r}^T, \tanh(\mathbf{r})^T]^T. \
\end{equation}
The $\tanh$ transformation exploits the saturation characteristics of reservoir states, sharpening class boundaries. This expansion enables the ridge regression to capture nonlinear decision boundaries while maintaining the computational benefits of closed-form training.
\item Support Vector Machine (SVM): A support vector classifier with an RBF kernel using a regularization parameter $C$ set to 10.0 and a kernel coefficient $\gamma$ determined by the scale heuristic. This model aims to maximize the margin for robust classification.
\item Random Forest (RF): Ensemble of 300 decision trees. This model handles high-dimensional features and provides robustness to outliers.
\end{itemize}

\section{Experimental Setup}

\subsection{Datasets}
We evaluate our method on two widely used benchmark datasets for radar-based HGR, namely the Soli dataset \cite{Lien2016Soli} and the Dop-NET dataset \cite{RitchieDopNet}.

\begin{figure}[t]
\centering
\includegraphics[width=\columnwidth]{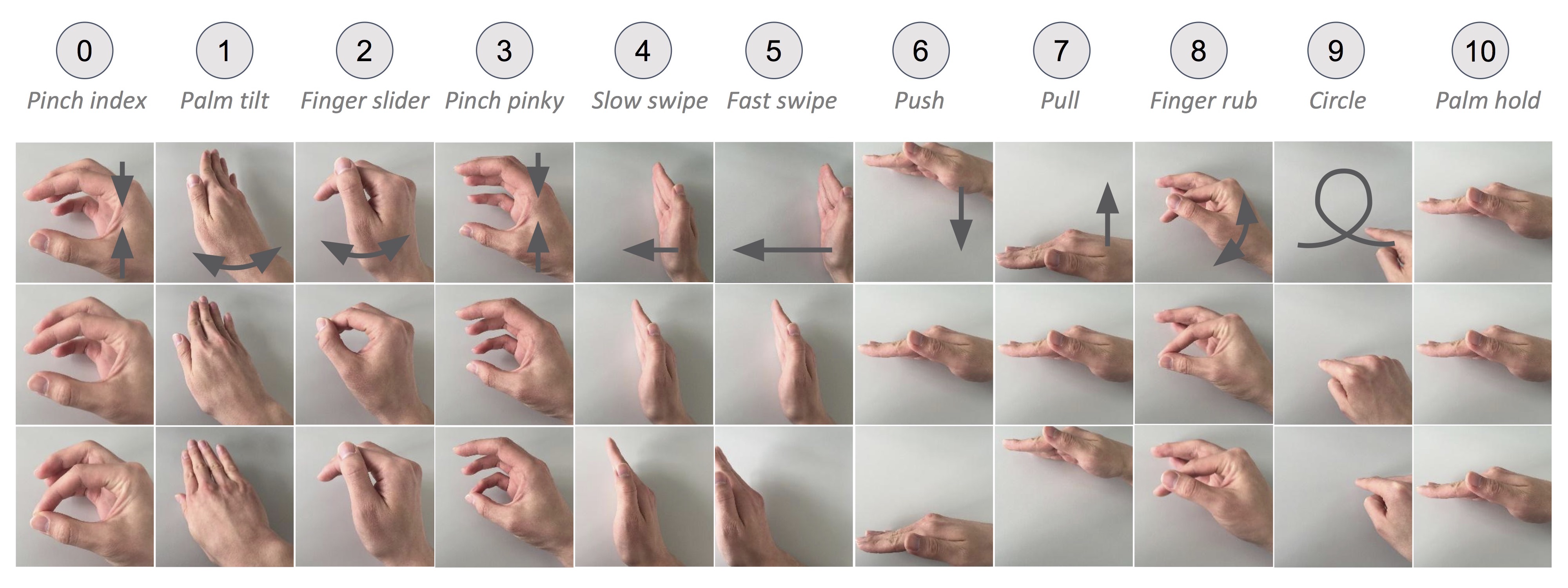}
\caption{Set of hand gestures included in the Soli dataset \cite{wang2016interacting}:
(0) Pinch index,
(1) Palm tilt,
(2) Finger slider,
(3) Pinch pinky,
(4) Slow swipe,
(5) Fast swipe,
(6) Push,
(7) Pull,
(8) Finger rub,
(9) Circle and
(10) Palm hold. Sample images from the deep-soli GitHub repository (https://github.com/simonwsw/deep-soli), licensed under MIT License.}
\label{fig:soli_gestures}
\end{figure}

The Soli dataset was collected using a 60GHz FMCW millmeter-wave radar module with one transmit and four receive antennas.
The dataset comprises 2,750 samples spanning 11 gesture classes.
Fig. \ref{fig:soli_gestures} illustrates the set of hand gestures included in the Soli dataset.
These gestures represent a diverse range of hand movements relevant to HCI applications, from fine finger manipulations to gross hand motions. 
Ten subjects participated in data collection, each performing 25 repetitions of each gesture across multiple sessions.
Table \ref{tab:soli_samples} shows the number of samples for each subject and each gesture.
\begin{table}[t]
\centering
\caption{Set of hand gestures included in the Soli dataset}
\label{tab:soli_samples}
\begin{tabular}{l|ccccc|c}
\hline
Person
& (0) & (1) & $\cdots$ & (9) & (10)
& Total \\
\hline
S1 & 25 & 25 & $\cdots$ & 25 & 25 & 275 \\
S2 & 25 & 25 & $\cdots$ & 25 & 25 & 275 \\
$\vdots$ & $\vdots$ & $\vdots$ & $\vdots$ & $\vdots$ & $\vdots$ & $\vdots$ \\
S9 & 25 & 25 & $\cdots$ & 25 & 25 & 275 \\
S10& 25 & 25 & $\cdots$ & 25 & 25 & 275 \\
\hline
Total
& 250 & 250 & $\cdots$ & 250 & 250
& 2750 \\
\hline
\end{tabular}
\end{table}
Each frame produces a $32 {\times} 32$ complex-valued RDM from each of the four receive channels, and the resulting data sequences are of variable length, ranging from 28 to 145 time steps. 

\begin{figure}[t]
\centering
\includegraphics[width=\columnwidth]{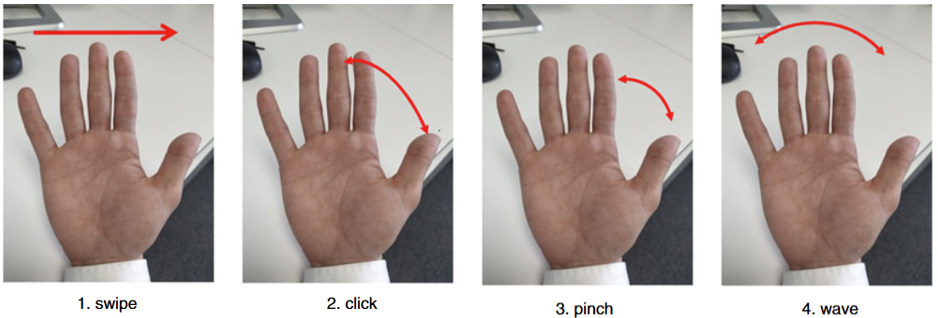}
\caption{Set of hand gestures included in the Dop-NET dataset \cite{RitchieDopNet}. Figure reproduced with permission from John Wiley and Sons.}
\label{fig:Dop_gestures}
\end{figure}

The Dop-NET dataset consists of radar-based hand gesture recordings collected using a 24 GHz FMCW millimeter-wave radar system equipped with one transmit antenna and two receive antennas. However, only the MDMs acquired from a single receive antenna are included in the dataset. The dataset comprises multiple gesture categories commonly used in micro-Doppler-based HGR.
Fig. \ref{fig:Dop_gestures} illustrates the set of hand gestures included in the Dop-NET dataset.
Six participants contributed to the dataset, with each subject performing repeated executions of each gesture across several recording sessions. Although a test set is provided, it is unlabeled. Therefore, only the training dataset are used in our experiments.
Table \ref{tab:training_samples} shows the number of training samples for each subject and each gesture.
\begin{table}[t]
\centering
\caption{Set of hand gestures included in the Dop-NET dataset}
\label{tab:training_samples}
\begin{tabular}{l|rrrr|r}
\hline
Person & Swipe & Click & Pinch & Wave & Total \\
\hline
A & 64 & 105 & 98 & 56 & 323 \\
B & 72 & 105 & 116 & 112 & 405 \\
C & 80 & 137 & 132 & 85 & 434 \\
D & 71 & 93 & 112 & 70 & 346 \\
E & 91 & 144 & 98 & 56 & 389 \\
F & 101 & 208 & 140 & 87 & 536 \\
\hline

Total & 479 & 792 & 696 & 466 & 2433 \\
\hline
\end{tabular}
\end{table}
Each gesture exhibits a distinct temporal duration, reflected in the varying lengths of the time axis, which range from 43 to 540 time steps, while the Doppler dimension is fixed to 800 bins. 

\subsection{ESN Configuration}
\label{sec:ESN}
For the Soli dataset, both a single-reservoir model and a multi-reservoir model are evaluated to enable a fair comparison between the two architectures.
As the Dop-NET dataset involves only a single receive antenna, a single-reservoir ESN architecture is used.
Table \ref{tab:params} summarizes the ESN hyperparameters.
Hyperparameter optimization was performed using Gaussian process–based Bayesian optimization implemented in scikit-optimize (skopt) via the gp\_minimize function \cite{scikit-optimize}.
\begin{table}[t]
  \centering
  \caption{ESN Hyperparameter Configuration}
  \label{tab:params}
  \small
  \begin{tabular}{lccc}
    \toprule
    Parameter
    & \makecell{Single-Res.\\(Soli)}
    & \makecell{Multi-Res.\\(Soli)}
    & \makecell{Single-Res.\\(Dop-NET)} \\
    \midrule
    Reservoir nodes & 500 & 50 (each) & 1000 \\
    Spectral radius & 0.95 & 0.95 & 1.1 \\
    Input scaling   & 0.1 & 1.0 & 0.4 \\
    Density         & 0.3 & 0.9 & 0.05 \\
    Leaking rate    & 0.01 & 0.0263 & 0.05 \\
    \bottomrule
  \end{tabular}
\end{table}

\subsection{Evaluation Protocols}
Classification performance is evaluated using accuracy as the primary metric. In addition to recognition accuracy, computational efficiency is assessed by measuring the execution time, including both training time and inference time.

We employ four evaluation protocols to comprehensively assess model performance, as described below.

\begin{enumerate}
\item 50:50 Holdout Split: Random partition into equal training and test sets. 
This setting provides a baseline performance estimate.

\item 10-Fold Cross-Validation: Stratified random partitioning into 10 folds with rotation. The results are given by mean accuracy and standard deviation across folds.

\item Leave-One-Subject-Out: Each fold holds out one subject's data for testing, and training is performed on the remaining subjects. This stringent protocol evaluates generalization to completely unseen individuals, reflecting practical deployment scenarios where the system must work for new users without retraining.

\item Leave-One-Session-Out: For each subject, the data are divided into training and test sets in a 50:50 ratio based on recording sessions, with each split used in turn for training and evaluation. This protocol evaluates temporal generalization and robustness to within-subject variations across recording sessions.
\end{enumerate}

\section{Results and Discussion}

\subsection{Classification Performance (50:50 Holdout Split And 10-Fold Cross Validation)}
Table~\ref{tab:results} presents classification accuracy across reservoir configurations and readout methods in all the evaluation protocols. 
Hereafter, we denote single-reservoir and multi-resevoir models as SR and MR, respectively.
We extend this notation to include the readout classifier name (e.g., SR-SVM denotes a single-reservoir model with an SVM classifier).

\begin{table*}[!b]
\centering
\caption{Classification Accuracy (\%) on Soli and Dop-NET Datasets}
\label{tab:results}
\resizebox{\textwidth}{!}{%
\begin{tabular}{lccccccc|ccc}
\toprule
& \multicolumn{3}{c}{Single-Res. (Soli)} 
& \multicolumn{4}{c}{Multi-Res. (Soli)} 
& \multicolumn{3}{c}{Single-Res. (Dop-NET)} \\
\cmidrule(lr){2-4} \cmidrule(lr){5-8} \cmidrule(lr){9-11}
Evaluation 
& RR$_{\mathrm{L}}$ & SVM & RF
& RR$_{\mathrm{L}}$ & RR$_{\mathrm{N}}$ & SVM & RF 
& RR$_{\mathrm{L}}$ & SVM & RF \\
\midrule
50:50 split
  & 92.90 & 96.80 & 94.70
  & 97.89 & \textbf{98.84} & 97.09 & 96.95
  & 91.62 & 94.74 & 91.21 \\
10-fold CV
  & 93.96$\pm$1.29 & 97.27$\pm$1.11 & 95.42$\pm$1.65
  & 98.07$\pm$0.67 & \textbf{98.73$\pm$0.69} & 97.53$\pm$0.90 & 96.84$\pm$1.08
  & 92.44$\pm$2.13 & 95.73$\pm$1.37 & 92.81$\pm$1.71 \\
Subject-Out
  & 85.45$\pm$8.37 & 89.71$\pm$6.59 & 82.00$\pm$8.98
  & 90.51$\pm$6.32 & \textbf{92.62$\pm$5.66}  & 90.65$\pm$7.00 & 87.20$\pm$8.75
  & 64.12$\pm$9.12 & 61.91$\pm$10.01 & 62.56$\pm$11.24 \\
Session-Out
  & 96.59$\pm$1.74 & 98.48$\pm$1.35 & 97.73$\pm$1.56
  & 97.34$\pm$2.52 & 97.20$\pm$2.58 &\textbf{98.51$\pm$1.31} & 97.97$\pm$1.63
  & 98.39$\pm$1.31 & 97.99$\pm$1.64 & 95.68$\pm$3.67 \\
\bottomrule
\end{tabular}
}
\end{table*}

For the Soli dataset, the MR model consistently outperforms the SR model across all protocols and classifiers.
To achieve the highest accuracy, we apply the RR$_{\mathrm{N}}$ readout only to the generally better-performing multi-reservoir models.
The MR-RR$_{\mathrm{N}}$ achieves the highest accuracy of 98.84\% on holdout split, representing a 0.95 percentage point improvement over MR-RR$_{\mathrm{L}}$ at 97.89\%, and a 5.94 percentage point improvement over SR-RR$_{\mathrm{L}}$ at 92.90\%.

Under 10-fold cross-validation, MR-RR$_{\mathrm{N}}$ achieves 98.73\%$\pm$0.69\%, with notably lower standard deviation indicating stable performance across folds.
For SVM and RF readouts, transitioning from a SR to a MR results in a performance improvement of approximately 0.3–2\%. 
These results imply that the MR configuration can mitigate interference between feature maps, leading to improved classification performance.

Notably, MR-RR$_{\mathrm{L}}$ achieves higher accuracy than MR-SVM and MR-RF. 
In contrast, in the SR, nonlinear classifiers achieve substantially higher accuracy than RR$_{\mathrm{L}}$.
These results suggest that introducing a proposed architecture enables linear classifiers to achieve strong performance.

On the Dop-NET dataset, the SR-SVM achieves a peak accuracy of 97.13\% under 10-fold cross-validation, while the average classification accuracy reaches 94.74\%. 
By mapping the 800 Doppler bins of the MDM into the reservoir, the feature representation is enriched, which facilitates more expressive representations.

\subsection{Class-Dependent Performance}
\begin{figure}[t]
\centering
\includegraphics[width=0.7\linewidth]{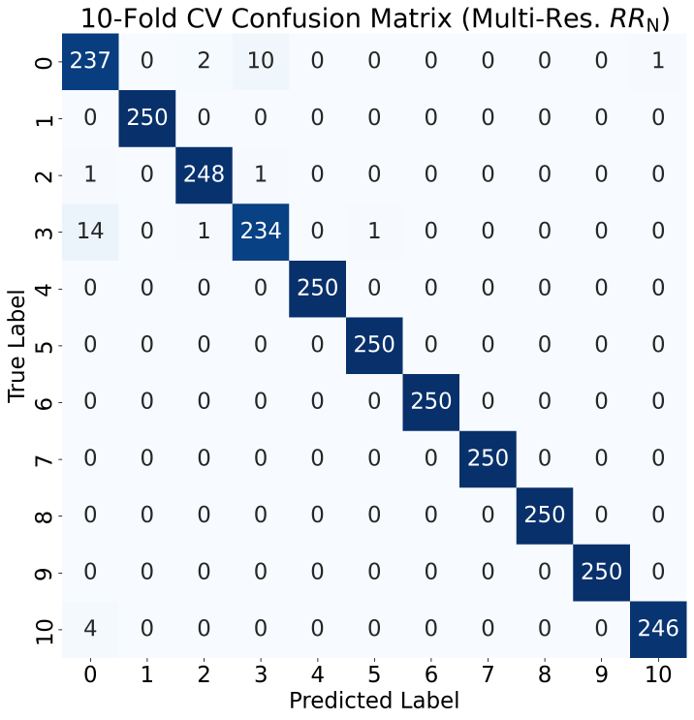}
\caption{Confusion matrix obtained using the multi-reservoir ESN with  RR$_{\mathrm{N}}$ for the Soli dataset.}
\label{fig:cm}
\end{figure}

Under the 10-fold cross validation protocol, we analyze a confusion matrix to reveal gesture-specific recognition characteristics. 
Fig. \ref{fig:cm} shows the confusion matrix for the Soli dataset using the MR-RR$_{\mathrm{N}}$, where each element represents the number of classified samples.
Gestures involving large-scale hand motions, including palm tilt, slow swipe, fast swipe, push, pull, finger rub, circle (Figs. 5(1), (4), (5), (6), (7), (8) and (9)) achieve perfect recognition accuracy.
These gestures generate strong and distinctive Doppler patterns, enabling reliable discrimination.

In contrast, two gesture classes exhibit relatively lower recognition performance, pinch index (Fig. 5(0)) and pinch pinky (Fig. 5(3)). 
Pinch index and pinch pinky consist of subtle finger-level movements with similar hand postures, which increases inter-class confusion. Furthermore, the range resolution of the 60 GHz radar, approximately four millimeters, is comparable to the scale of finger motions, limiting the separability of such fine-grained gestures.
However, since these samples are not misclassified as Palm Hold (Fig. 5(k)), the results suggest that the radar is capable of detecting subtle motions, although it may not sufficiently discriminate fine-grained differences among these gesture classes.

\begin{figure}[t]
\centering
\includegraphics[width=0.7\linewidth]{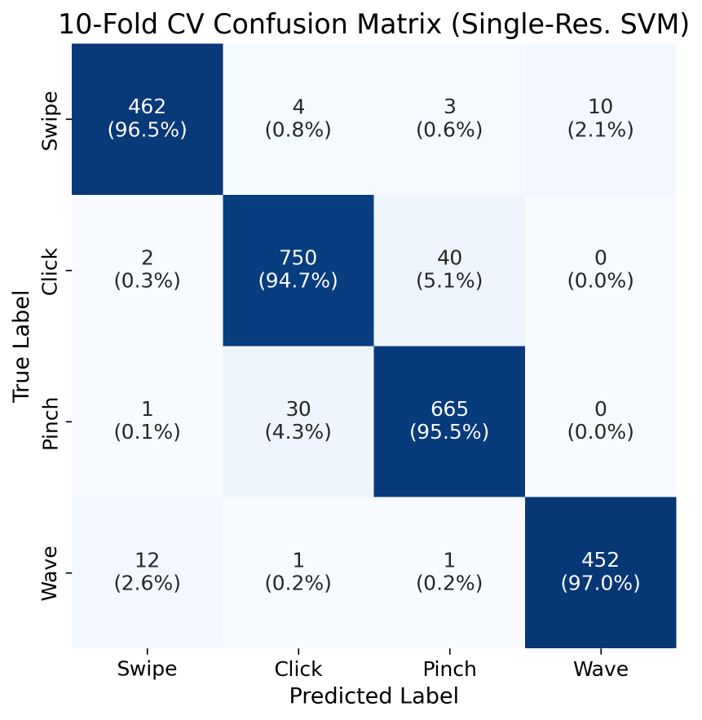}
\caption{Confusion matrix obtained using the single-reservoir SVM for the Dop-NET dataset.}
\label{fig:cm_dopnet}
\end{figure}

Fig. \ref{fig:cm_dopnet} shows the confusion matrix for the Dop-NET dataset using the SR-SVM. To account for class imbalance, values are expressed as accuracy rates in percentage. Each cell reports both the number of samples classified into the corresponding class and the associated accuracy.

The wave gesture achieves the highest recognition accuracy among all classes. As wave is a repetitive action, it consists of relatively longer time steps compared to other gestures. This likely enables the reservoir to capture richer temporal patterns, which contribute to the improved classification performance.
The recognition accuracy for click and pinch gesture is low compared to other classes. 
A possible cause is that the Dop-NET dataset contains samples with high noise levels. 
Since our method does not perform explicit preprocessing or noise removal, these noise components likely degrade the quality of the extracted feature representations. 
In particular, when two gesture classes exhibit highly similar motion patterns, additional reflections from the wrist or other body parts can be captured as noise, making it difficult to reliably distinguish between them and thereby adversely affecting classification performance.

\subsection{Cross-Subject And Cross-Session Performance}
Under the leave-one-subject-out evaluation, MR-RR$_{\mathrm{N}}$ achieves an accuracy of 92.62\%  on the Soli dataset, accompanied by an increased standard deviation compared to 10-fold cross-validation, indicating substantial inter-subject variability. 
While recognition rates exceeds 95\% for certain subjects, they drop below 85\% for others. 
In contrast, under the leave-one-session-out evaluation, MR-SVM achieves a substantially higher accuracy of 98.51\%.

Under cross-subject evaluation, the Dop-NET dataset yields considerably lower performance, with average accuracies around 60\%. 
To date, no model has been reported to exceed 70\% average accuracy under split cross-validation, underscoring the substantial inter-subject variability and the difficulty of achieving robust generalization. In contrast, under cross-session evaluation, the Dop-NET dataset attains an accuracy of 97.99\% using SR-SVM, indicating substantially higher performance than in the cross-subject setting.

These results suggest that gesture representations remain highly consistent across sessions for the same individual, making within-subject generalization considerably easier than cross-subject generalization.
This observation implies that not only micro-Doppler characteristics but also DTM and RTM are strongly influenced by individual-specific motion traits, such as hand size, execution style, and movement speed.
Consequently, mitigating inter-subject variability will require modeling approaches that explicitly account for such individual-dependent variations. 

\subsection{Execution Time}
Table~\ref{tab:time} summarizes training and inference times measured on a standard PC equipped with an AMD Ryzen 9 9950X 16-core (32-thread) processor and 128 GB RAM.
\begin{table}[t]
\centering
\caption{Computational Time Analysis}
\label{tab:time}

\begin{tabular}{llcc}
\toprule
Dataset & Configuration & Training [s] & Inference [ms/sample] \\
\midrule
\textbf{Soli} & Single-Res. RR$_{\mathrm{L}}$ & 2.91& 2.04\\
& Single-Res. SVM & 2.92& 2.11\\
& Single-Res. RF & 3.14& 2.06\\
\cmidrule{2-4}
& Multi-Res. RR$_{\mathrm{L}}$ & 1.80 & 1.32\\
& Multi-Res. RR$_{\mathrm{N}}$ & 1.88 & 1.36\\
& Multi-Res. SVM & 2.23& 1.65\\
& Multi-Res. RF & 2.43& 1.63\\
\midrule
\textbf{Dop-NET} & Single-Res. RR$_{\mathrm{L}}$ & 7.39 & 6.36 \\
& Single-Res. SVM & 7.49 & 6.45 \\
& Single-Res. RF & 7.64 & 6.39 \\
\bottomrule
\end{tabular}

\end{table}
On the Soli dataset, per-sample inference times range from approximately 1.3 to 2.1 ms, and training completes within 3 seconds for all configurations under the 50:50 hold-out split. In contrast, for the Dop-NET dataset, inference times are approximately 6 ms per sample, and training times are around 7 seconds.

Based on the Soli results, the MR demonstrates advantages over the SR in terms of both training and inference time. Since these times depend on the dimensionality of the reservoir states used for readout, which corresponds to the total number of reservoir nodes, setting this dimensionality smaller than that of the SR enables the efficiency gains achieved by the MR.

\subsection{Comparison with Other Methods}
Table \ref{tab:comparison} compares our method with existing approaches under comparable evaluation protocols.
\begin{table}[t]
\centering
\caption{Comparison with Existing Methods on Soli and Dop-NET Datasets}
\label{tab:comparison}

\begin{tabular}{lcccc}
\toprule
Method & \makecell{Best \\ Accuracy} & Dataset & Data Split & Reference \\
\midrule
CNN + RNN                & 77.71\%   & Soli    & 50:50 & \cite{wang2016interacting} \\
CNN + LSTM               & 87.17\%   & Soli    & 50:50 & \cite{wang2016interacting} \\
ResNet-18 + LSTM         & 92.55\%   & Soli    & 60:40 & \cite{Zhang2022ResNetLSTM} \\
LSM                      & 98.02\%   & Soli    & 50:50 & \cite{tsang2021radar} \\
ESN (MR-RR$_{\mathrm{N}}$)      & 98.84\%   & Soli    & 50:50 & this work \\
\midrule
LSM              & 98.91\%   & Dop-NET & 50:50 & \cite{tsang2021radar} \\
K-Nearest Neighbor       & 89.00\%   & Dop-NET & 90:10    & \cite{Ritchie2019MicroDoppler} \\
ESN (SR-SVM)    & 94.74\%   & Dop-NET & 50:50 & this work \\
\bottomrule
\end{tabular}

\end{table}

On the Soli dataset, our approach outperforms all compared methods. Notably, the proposed architecture achieves an accuracy that is 0.82 percentage points higher than the best score obtained by LSM, which similarly utilizes RC but necessitates complex spike encoding preprocessing. When compared to deep learning models such as CNN-LSTM and ResNet-LSTM, our multi-reservoir ESN approach provides a performance gain ranging from 6 to 11\%. This superior classification accuracy is achieved while requiring orders of magnitude fewer trainable parameters, highlighting the significant efficiency of the proposed model.

For the Dop-NET dataset, the obtained results indicate that RC-based models, such as ESN and LSM, are well suited to this task.
The temporal dynamics and Doppler-centric representations contained in Dop-NET can be effectively captured by these models without relying on complex feature engineering or deep architectures.
This observation suggests that the intrinsic memory and nonlinear processing capabilities of reservoir computing provide a natural match to the characteristics of the Dop-NET dataset.

\section{Conclusion}
This work proposed an ESN-based approach with a parallel multi-reservoir architecture for hand gesture recognition using Doppler radar signals. To leverage different types of feature maps extracted from raw radar signals and prevent their interference within the reservoir, we integrated those information at the stage of readout computation. The experimental results with the Soli dataset demonstrated that the proposed model outperforms the single-reservoir counterpart and other tested models.

Based on the results of this study, several directions for future work can be considered.
First, the expansion of recognition targets is an important direction. While this study focused on hand gestures, it can also be applied to non-contact monitoring such as gait analysis and identification of activities of daily living. Validating the dynamic pattern recognition capability of ESN for these diverse targets represents an extremely meaningful endeavor.
Second, improving robustness in real-world environments is required for practical applications. Separating the target signal becomes challenging in environments where multiple people are present simultaneously or where dynamic background noise exists.
Third, model implementation and optimization for resource-constrained edge devices are key challenges. ESN is highly suitable for embedded systems due to its extremely low computational load for both learning and inference. Running this method in real-time on hardware and evaluating its power efficiency and responsiveness represents a crucial step toward practical implementation.


\begin{thebibliography}{00}

\bibitem{molchanov2016online} P. Molchanov et al., ``Online detection and classification of dynamic hand gestures with recurrent 3D convolutional neural networks,'' in Proc. IEEE CVPR, 2016, pp. 4207--4215.
\bibitem{ahuja2021vid2doppler} K. Ahuja et al., ``Vid2Doppler: Synthesizing Doppler radar data from videos for training privacy-preserving activity recognition,'' in Proc. ACM CHI, 2021, pp. 1--10.
\bibitem{Ahmed2021hand} S. Ahmed et al., ``Hand gesture recognition using radar sensors for human-computer-interaction: A review,'' Remote Sens., vol. 13, no. 3, p. 527, 2021.
\bibitem{Soumya2023Review} A. Soumya et al., ``Recent advances in mmWave radar-based sensing for indoor applications,'' Sensors, vol. 23, no. 4, p. 2241, 2023.
\bibitem{jankiraman2018fmcw} M. Jankiraman, FMCW Radar Design. Artech House, 2018.
\bibitem{Skaria2023MmWaveReview} S. Skaria et al., ``Gesture recognition using mmWave radar: A survey,'' IEEE Sensors J., vol. 23, no. 8, pp. 8320--8335, 2023.
\bibitem{jaeger2001echo} H. Jaeger, ``The echo state approach to analysing and training recurrent neural networks,'' GMD Tech. Rep. 148, 2001.
\bibitem{tanaka2019reservoir} G. Tanaka et al., ``Recent advances in physical reservoir computing: A review,'' Neural Netw., vol. 115, pp. 100--123, 2019.
\bibitem{gallicchio2017deep}
C. Gallicchio, A. Micheli, and L. Pedrelli, 
``Deep reservoir computing: A critical experimental analysis,'' 
\textit{Neurocomputing}, vol. 268, pp. 87--99, 2017.
\bibitem{li2023multi}
Z. Li, Y. Liu, and G. Tanaka, 
``Multi-reservoir echo state networks with Hodrick--Prescott filter for nonlinear time-series prediction,'' 
\textit{Applied Soft Computing}, vol. 135, p. 110021, 2023.
\bibitem{wang2016interacting} S. Wang et al., ``Interacting with Soli: Exploring fine-grained dynamic gesture recognition in the radio-frequency spectrum,'' in Proc. ACM UIST, 2016, pp. 851--860.
\bibitem{RitchieDopNet}
M. Ritchie, R. Capraru, and F. Fioranelli, ``Dop-NET: A micro-Doppler radar data challenge,'' in \emph{Electronics Letters}, vol. 56, no. 11, pp. 568--570, 2020.
\bibitem{Lien2016Soli} J. Lien et al., ``Soli: Ubiquitous gesture sensing with millimeter wave radar,'' ACM Trans. Graph., vol. 35, no. 4, pp. 1--19, 2016.
\bibitem{Zhang2022ResNetLSTM} Y. Zhang et al., ``Dynamic hand gesture recognition using ResNet-LSTM network based on radar sensor,'' Sensors, vol. 22, no. 5, p. 1852, 2022.
\bibitem{yan2024emerging}
M. Yan, C. Huang, P. Bienstman, P. Tino, W. Lin, and J. Sun, 
``Emerging opportunities and challenges for the future of reservoir computing,'' 
\textit{Nature Communications}, vol. 15, no. 1, p. 2056, 2024.
\bibitem{tsang2021radar} I. Tsang et al., ``Radar-based hand gesture recognition using spiking neural networks,'' Electronics, vol. 10, no. 12, p. 1405, 2021.
\bibitem{Maass2002RealTime}
W. Maass, T. Natschläger, and H. Markram, ``Real-time computing without stable states: A new framework for neural computation based on perturbations,'' \emph{Neural Computation}, vol.~14, no.~11, pp.~2531--2560, 2002.

\bibitem{jaeger2007optimization}
H. Jaeger, M. Luko\v{s}evi\v{c}ius, D. Popovici, and U. Siewert,
``Optimization and applications of echo state networks with leaky-integrator neurons,''
\textit{Neural Networks}, vol. 20, no. 3, pp. 335--352, 2007.
\bibitem{scikit-optimize}
The scikit-optimize developers,
``scikit-optimize,''
[Online]. Available: https://scikit-optimize.github.io/. Accessed Jan. 28, 2026.

\bibitem{Ritchie2019MicroDoppler}
M. Ritchie and A. M. Jones, ``Micro-Doppler gesture recognition using Doppler, time and range based features,'' in \emph{Proc. IEEE Radar Conf. (RadarConf)}, 2019, pp.~1--6.






\end{thebibliography}
\end{document}